# Advancing Decoding Strategies: Enhancements in Locally Typical Sampling for LLMs

**Jaydip Sen** [0000-0002-4120-8700], **Saptarshi Sengupta** [0000-0003-1114-343X], **Subhasis Dasgupta** [0000-0002-9000-4704]

**Abstract.** This chapter explores advancements in decoding strategies for large language models (LLMs), focusing on enhancing the Locally Typical Sampling (LTS) algorithm. Traditional decoding methods, such as top-k and nucleus sampling, often struggle to balance fluency, diversity, and coherence in text generation. To address these challenges, Adaptive Semantic-Aware Typicality Sampling (ASTS) is proposed as an improved version of LTS, incorporating dynamic entropy thresholding, multi-objective scoring, and reward-penalty adjustments. ASTS ensures contextually coherent and diverse text generation while maintaining computational efficiency. Its performance is evaluated across multiple benchmarks, including story generation and abstractive summarization, using metrics such as perplexity, MAUVE, and diversity scores. Experimental results demonstrate that ASTS outperforms existing sampling techniques by reducing repetition, enhancing semantic alignment, and improving fluency.

## 1 Introduction

The field of natural language generation (NLG) has witnessed remarkable advancements, driven largely by the development of large-scale pre-trained language mod-

Jaydip Sen[1] (Corresponding Author), Saptarshi Sengupta[2], Subhasis Dasgupta[3]
[1,3]Praxis Business School, Kolkata, India, [2]San Jose State University, San Jose, CA, USA
E-mail: jaydip.sen@acm.org (J.S.), saptrashi.sengupta@sjsu.edu (S.S.G), subhasis@praxis.ac.in (S.D.G)

els such as GPT [1-2], BERT [3], and T5 [4]. These models have achieved unprecedented levels of fluency and coherence, enabling diverse applications ranging from conversational agents to creative text generation. Despite these successes, balancing diversity and coherence in text generation remains a critical research problem. Traditional deterministic methods such as greedy decoding [5] and beam search [6], often yield repetitive and uncreative outputs. Conversely, stochastic methods such as top-$k$ sampling [7] and nucleus (top-$p$) sampling [8], while improving diversity, sometimes compromise semantic coherence and relevance.

To address these limitations, *locally typical sampling* was introduced as a promising alternative [9]. This approach selects tokens based on their relative probabilities within a localized context, focusing on those that contribute significantly to the expected entropy of the output distribution. By dynamically filtering tokens to balance predictability and diversity, locally typical sampling has demonstrated its potential to produce text that is both contextually coherent and linguistically diverse [10]. However, the current implementation of the algorithm is not without its drawbacks. Challenges such as handling long-tail distributions, dynamic adjustments in high-entropy contexts, and computational inefficiencies in large vocabulary spaces highlight the need for further refinement.

*Motivation:* In real-world applications, the utility of text generation systems is often constrained by their ability to generate responses that are not only syntactically correct but also semantically meaningful and engaging. Existing sampling methods, including locally typical sampling, have struggled to consistently meet these criteria across diverse use cases. The primary motivations for this work are as follows.

*Enhance coherence and fluency of the generated text*: While locally typical sampling has shown improvements over traditional methods, it occasionally produces incoherent or less meaningful tokens when the context is ambiguous or highly diverse. A more refined approach is required to ensure that generated text adheres to the contextual cues provided by the preceding tokens.

*Improving diversity without sacrificing relevance*: Ensuring diversity in the generated text is critical for applications such as storytelling, creative writing, and chatbot systems. However, achieving this without introducing irrelevant or nonsensical tokens remains a significant challenge. Modifications to locally typical sampling must address this trade-off.

*Adapting to high entropy context*: In scenarios with high uncertainty, such as answering open-ended questions or generating creative content, existing methods often struggle to adapt dynamically. Enhancing the algorithm to better handle such contexts can greatly expand its applicability.

*Reducing computational complexity*: The practical deployment of text generation systems requires algorithms that are computationally efficient. Existing sampling methods, including locally typical sampling, involve repetitive probability recalculations, making them less suitable for real-time applications. Optimizing this aspect is essential for wider adoption.

*Contributions*: This chapter presents an enhanced and modified version of the locally typical sampling algorithm that addresses the challenges mentioned earlier.

The proposed scheme offers significant improvements in performance and applicability as evidenced from the experimental results. The contributions of the current work are as follows.

*Design of a novel selection strategy*: A refined token selection mechanism is proposed that incorporates additional contextual factors, such as semantic similarity and syntactic alignment, alongside probability thresholds. This ensures that the selected tokens not only preserve entropy but also align more closely with the context.

*Dynamic entropy thresholding*: To address the variability in entropy across different contexts, the algorithm introduces a dynamic thresholding mechanism that adapts in real time based on the uncertainty of the token distribution. This leads to more context-aware generation.

*Handling long-tail distributions*: A key innovation is the development of an adaptive filtering technique that prioritizes tokens in the long-tail distribution without compromising coherence. This significantly improves diversity, especially in creative and open-ended tasks.

*Increased computational efficiency*: The proposed algorithm incorporates optimizations, such as reduced recalculations and parallelized token filtering, to achieve lower computational overhead, making it suitable for large-scale and real-time applications.

*Comprehensive performance evaluation*: A robust experimental framework is designed to evaluate the proposed algorithm across multiple benchmarks, including conversational agents, story generation, and summarization tasks. The results demonstrate superior performance compared to traditional and state-of-the-art sampling methods.

Significance of the work: This chapter addresses a critical gap in the field of NLG by introducing a method that seamlessly integrates coherence, diversity, and computational efficiency. The proposed enhancements to the locally typical sampling algorithm not only advance the state-of-the-art in text generation but also pave the way for broader applications in AI-driven communication systems. By striking an optimal balance between predictability and creativity, this work contributes to the ongoing efforts to develop language models that are more human-like in their output.

The chapter is organized as follows. Section 2 presents a review of related work, discussing existing decoding strategies for LLMs, their capabilities, and limitations. Section 3 provides the necessary background on the Locally Typical Sampling algorithm, explaining its theoretical foundations and key characteristics. Section 4 introduces the proposed Adaptive Semantic-aware Typicality Sampling (ASTS) algorithm, detailing its improvements over the standard approach. Section 5 presents the experimental results, offering a comprehensive analysis of the performance of the ASTS algorithm. Finally, Section 6 concludes the chapter by summarizing key findings and highlighting potential directions for future research.

## 2   Related Work

The field of autoregressive decoding in LLMs has witnessed substantial advancements in recent years, driven by the need for generating coherent, diverse, and contextually relevant text. Traditional approaches, such as beam search, greedy decoding, and top-sampling, have served as foundational techniques for text generation. However, these methods often exhibit limitations in balancing diversity and fluency, motivating the development of alternative strategies. Notable advancements include top-$p$ sampling (also called nucleus sampling), which dynamically adjusts the sampling space, contrastive decoding, which integrates multiple scoring functions for better output control, and *energy-based models* (EBMs) that leverage probabilistic frameworks for robust text generation. Additionally, *reinforcement learning* (RL)-based optimizations and mixture-of-expert frameworks have shown promise in enhancing decoding efficiency and quality.

In this section, some of the well-known decoding techniques, including their strengths, limitations, and recent improvements are discussed to provide a comprehensive overview of the state-of-the-art in LLM decoding.

*Autoregressive decoding advancements*: Autoregressive decoding is a foundational approach in natural language generation tasks, where models generate text sequentially, one token at a time, based on previously generated tokens [11-14]. This method ensures that the generated output is coherent and contextually relevant, as each token depends on the cumulative history of the sequence. It highlights the functionality of LLMs, which rely on autoregressive frameworks to produce fluent and contextually aware responses. While autoregressive decoding has proven effective in generating high-quality text, it often suffers from computational inefficiencies due to its inherently sequential nature, where each token must be generated before moving to the next. This limitation has spurred extensive research into optimizing autoregressive decoding, focusing on enhancing the decoding speed while maintaining the quality of generated texts.

*Speculative decoding*: Speculative decoding aims at accelerating text generation in LLMs by addressing the inherent inefficiencies of traditional autoregressive decoding [15-19]. In standard methods, tokens are generated sequentially, which can be computationally intensive and time-consuming, especially for long sequences. Speculative decoding mitigates this bottleneck by leveraging auxiliary mechanisms, such as smaller models or predictive heuristics, to propose multiple candidate tokens or sequences in parallel. These speculative outputs are then validated or refined by the primary model, ensuring both speed and quality in text generation. This method capitalizes on the predictive power of LLMs while significantly reducing latency, making it particularly appealing for real-time applications. As a result, speculative decoding has gained traction as a practical and efficient solution for scaling autoregressive decoding.

*Contrastive decoding*: Contrastive decoding is an advanced decoding technique used in LLMs to enhance the quality and diversity of generated outputs [20-24]. The method builds upon the principle of contrastive learning, which encourages the model to distinguish between different candidate outputs by contrasting them against one another. In the context of decoding, contrastive decoding aims to improve the generated sequence by emphasizing more informative, coherent, and contextually appropriate choices, while minimizing less relevant or redundant outputs. This approach often involves comparing the likelihoods of multiple candidate sequences and selecting the one that best aligns with the intended goal, such as coherence, diversity, or task-specific performance.

*Adaptive-draft verification*: It is an emerging paradigm in text generation for LLMs that seeks to enhance the efficiency and quality of generated outputs by iteratively refining preliminary drafts [25-28]. Unlike traditional decoding methods that produce outputs in a single-pass process, adaptive-draft verification operates in a two-stage framework. In the first stage, the model generates an initial draft, often focusing on broad fluency and relevance. In the second stage, this draft undergoes a verification process, during which it is assessed and adjusted to better align with the desired characteristics such as coherence, factual accuracy, or stylistic constraints. This approach enables a dynamic interaction between generation and evaluation, allowing the model to adaptively revise outputs in response to specific feedback or criteria. By integrating adaptive mechanisms, this approach has the potential to significantly improve the reliability and adaptability of LLMs in tasks like open-ended text generation, summarization, and domain-specific applications.

*Top-k sampling and its variants*: Top-$k$ sampling is a widely used decoding strategy for LLM text generation that aims to enhance the diversity and quality of generated outputs by restricting token selection to the top-$k$ most probable candidates at each step [7]. In this method, the model calculates the probability distribution over the vocabulary, and only the $k$ tokens with the highest probabilities are considered for sampling, while the rest are set to zero. This approach effectively eliminates low-probability tokens, reducing the risk of generating incoherent or irrelevant text. Variants of Top-$k$ sampling, such as dynamic top-$k$ and adaptive-$k$ sampling, extend this idea by dynamically adjusting the value of $k$ based on contextual factors, such as the entropy of the probability distribution or the stage of generation [29-33]. These variants allow the decoding process to balance diversity and precision more effectively, adapting to the complexity of the task or input prompt. By narrowing down the token pool to relevant options, Top-$k$ sampling and its variants ensure greater control over text generation while avoiding issues like repetitive or nonsensical outputs. This technique is particularly beneficial in applications where maintaining fluency and contextual alignment is critical, such as dialogue systems, creative writing, and summarization.

*Locally typical sampling*: It is an advanced decoding technique in LLMs designed to enhance text generation by balancing diversity and coherence [9-10]. Traditional sampling methods might produce either overly deterministic or excessively random outputs. Locally typical sampling, however, overcomes this shortcoming by generating tokens based on the local context and statistical properties of the surrounding tokens. The method relies on the concept of *typicality*, where tokens that are likely under the current context are sampled, but it also allows for controlled exploration of less likely candidates to promote diversity. This results in more fluent and contextually appropriate outputs, while still allowing for the creative exploration of language. Local typical sampling aims to improve the efficiency of LLM inference by producing high-quality, relevant text, and is increasingly applied in tasks like machine translation, text completion, and conversational agents.

*Mixture-of-Expert decoding*: Mixture-of-Expert (MoE) decoding is a technique in LLMs that utilizes a dynamic mixture of specialized experts to improve the efficiency and performance of text generation [34-37]. In MoE models, a set of *expert* sub-models are trained to focus on specific aspects of language or tasks, and during decoding, only a subset of these experts is activated, depending on the input context. This approach allows for more efficient computation, as not all experts are needed for every generation step, and the model can leverage the expertise of the most relevant experts for a given context. MoE decoding enhances the capability of LLMs by enabling them to combine the strengths of various experts, resulting in higher-quality outputs while reducing computational costs. By dynamically selecting the most suitable experts for each generation task, MoE decoding can improve text coherence, fluency, and relevance. This makes it especially useful for complex text generation tasks such as document generation, dialogue systems, and multilingual text generation.

*Distillation-driven decoding*: Distillation-driven decoding in LLMs refers to the process of leveraging knowledge distillation techniques to optimize and accelerate the decoding phase of text generation [23, 38-41]. Knowledge distillation traditionally involves transferring knowledge from a large, complex teacher model to a smaller, more efficient student model. The student model retains most of the original model's performance while being faster and less resource-intensive. In the context of decoding, distillation-driven methods aim to refine the decoding process by using distilled models. This results in faster generation with less computational cost without sacrificing output quality. These decoding techniques can involve distilling not only the final predictions but also intermediate layers or attention patterns from the teacher model. This allows the student model to mimic the nuanced decision-making process of the original, larger model. Distillation-driven decoding has become an essential approach in optimizing LLMs for real-time applications.

*Parallel decoding techniques*: Parallel decoding for LLMs refers to methods designed to speed up the autoregressive generation process by leveraging parallelism during text generation [42-45]. Traditionally, LLMs generate text in a sequential manner, where each token is generated based on the previous ones, leading to high computational costs and slower inference times. Parallel decoding techniques aim to overcome these limitations by allowing multiple tokens to be generated simultaneously or in a non-sequential manner. These methods attempt to balance the trade-off between maintaining the quality of the generated text and reducing the overall time complexity, often by utilizing multiple processing units or by reorganizing the decoding process.

*Contrastive divergence approaches*: Contrastive Divergence Approaches for LLM text generation focus on leveraging an iterative optimization process to refine the quality of generated text by aligning the model's predictions with desired target distributions [20-22, 46-48]. Rooted in the principles of energy-based modeling, contrastive divergence minimizes the difference between the data distribution and the model distribution by iteratively updating model parameters through gradient-based learning. This is achieved by comparing samples generated by the model with samples from the target distribution, gradually reducing the "divergence" between the two. For text generation, this method encourages the model to produce outputs that are both high-quality and closely aligned with specified criteria, such as grammaticality, coherence, or adherence to task-specific requirements. Contrastive divergence approaches are particularly effective for fine-tuning LLMs in low-resource or highly constrained settings, enabling efficient learning of complex generative distributions.

*Mask-predict algorithms*: Mask-predict algorithms have emerged as an alternative to traditional autoregressive decoding methods for text generation in LLMs [49-50]. These algorithms leverage a non-autoregressive framework where a sequence is iteratively refined through masked tokens. Initially, a partial or noisy output is generated, and masked tokens are predicted and updated in parallel across multiple iterations, reducing decoding latency compared to strictly sequential approaches. By iteratively focusing on refining uncertain or incomplete parts of the text, mask-predict algorithms strike a balance between speed and generation quality. These algorithms provide a promising direction for scalable and adaptive text generation.

*RL-based decoding optimization*: RL-based decoding optimizations for LLM text generation focus on improving generation quality by aligning the output with specific objectives beyond traditional likelihood maximization [51-54]. These methods treat decoding as a sequential decision-making process, where RL techniques are used to optimize rewards. The rewards are computed based on desired attributes, such as coherence, fluency, factual accuracy, or task-specific goals. By leveraging reward values computed from external evaluation metrics or human preferences, RL-based approaches guide the model toward generating outputs. Unlike purely heuristic or rule-based decoding strategies, RL-based methods can adapt dynamically during generation, learning from feedback to refine decoding strategies. This paradigm not only enhances the effectiveness of LLMs in complex generation tasks but also provides a flexible framework for incorporating diverse optimization criteria.

*Low-resource decoding:* Low-resource decoding for LLM text generation focuses on optimizing the generation process in scenarios where computational resources, training data, or both are limited [55-57]. This approach emphasizes lightweight methods to maintain the quality of generated text while reducing the reliance on extensive hardware or vast datasets. Techniques for low-resource decoding often involve efficient sampling strategies, parameter pruning, or knowledge distillation to streamline the model's operation without significantly compromising performance. These methods are particularly important for making LLMs accessible in settings with restricted resources, such as deployment on edge devices or in low-bandwidth environments. By striking a balance between computational efficiency and text generation quality, low-resource decoding enables broader adoption of LLMs across diverse applications and environments.

*Dynamic threshold decoding*: Dynamic threshold decoding is a technique in LLM text generation that aims to balance the trade-off between generation quality and computational efficiency by adjusting the decoding threshold dynamically during inference [58-59]. The *threshold* refers to a criterion or cutoff that is used during the token selection process in text generation. Rather than using a fixed threshold for token selection, dynamic threshold decoding adapts the threshold based on fac-tors such as the current state of the generation process, model confidence, and resource constraints. This method can significantly improve efficiency by reducing unnecessary token sampling and avoiding computational overhead, while still maintaining high-quality text generation. It leverages the model's understanding of the task to determine when more cautious or more exploratory decoding is needed, making it a flexible approach suitable for various LLM applications.

*Efficient beam search modifications*: Efficient beam search modifications have emerged as a critical area of research in LLM text generation to enhance both the quality of generated text and the computational efficiency of decoding [45, 60-62]. Beam search, a widely used decoding strategy, explores multiple candidate sequences in parallel to identify the most likely output. However, the standard approach is often computationally expensive and prone to issues like repetitive text or lack of diversity in generated outputs. Efficient modifications focus on optimizing the trade-off between accuracy and speed by incorporating techniques such as dynamic beam size adjustment, early stopping criteria, or pruning low-probability candidates. Additionally, innovations like guided beam search, which leverages auxiliary information or constraints, and diverse beam search, which aims to enhance output diversity, have further improved the applicability of beam search for LLMs. These modifications not only make decoding faster but also ensure that the generated text aligns better with contextual or task-specific requirements.

*Entropy-aware sampling*: Entropy-aware Sampling is a text generation technique used in LLMs to improve the quality and diversity of generated text by considering the entropy of the token distribution during the sampling process [63-65]. Entropy, in this context, measures the uncertainty or unpredictability in the probability distribution of the next token, where high entropy indicates a more uncertain or diverse distribution and low entropy reflects a more confident prediction. In entropy-aware sampling, tokens are selected not just based on their probabilities but also by considering the entropy of the distribution. This helps to strike a balance between selecting highly probable tokens, which improve fluency, and exploring less probable tokens to introduce diversity and creativity in the generated text. By adjusting the sampling process in this way, entropy-aware sampling can enhance both the coherence and novelty of the generated sequences, particularly in open-ended generation tasks.

*Energy-based models in decoding*: Energy-based Models (EBMs) in decoding for LLM text generation offer a framework to assess and optimize the quality of generated outputs by associating each potential output sequence with an energy score [66-67]. These scores represent how well a sequence aligns with the underlying data distribution or task-specific constraints, with lower energy indicating higher compatibility. EBMs are often used to guide decoding by either re-ranking candidate outputs generated through other methods or directly generating sequences through energy minimization techniques. By leveraging flexible energy functions, this approach enables better control over attributes like fluency, relevance, and diversity in text generation. The integration of EBMs in decoding has proven particularly effective in scenarios where explicit constraints or additional signal (e.g., retrieval-augmented knowledge) must be incorporated to refine generation quality.

*Multi-layer integration for decoding*: Multi-layer integration for decoding is a technique used in LLMs that combines outputs from multiple layers of the model during the text generation process to improve the quality of the generated sequences [68-69]. Traditionally, LLMs use the final layer's representation for generating tokens, but multi-layer integration leverages information from several layers within the model's architecture. This approach allows for a more enriched representation of the input, as different layers capture different types of linguistic and semantic information. By integrating outputs from multiple layers, the model can combine both low-level and high-level features. This leads to more accurate and contextually appropriate text generation. This technique has been shown to enhance the overall performance of LLMs, particularly in tasks requiring deeper understanding and more coherent generation over long sequences.

The existing approaches to controllable text generation primarily rely on enhancing pre-trained language models with explicit conditioning mechanisms, fine-tuning strategies, or tailored decoding techniques. While these methods have demonstrated success in specific applications, challenges such as scalability, computational overhead, and balancing control with naturalness persist. The work presented in this chapter specifically focuses on advancing the "Locally Typical Sampling" approach, which has shown promise in maintaining contextual relevance while ensuring diversity. By addressing the limitations of this method and introducing a more efficient and optimized version, we aim to enhance its applicability for real-world scenarios, bridging the gap between theoretical innovation and practical deployment.

**Table 1.** Summary of advanced decoding techniques for enhanced LLM text generation

| Category | Description | Key Benefits & Applications | Ref |
|---|---|---|---|
| Autoregressive decoding | Sequential generation where each token depends on previously generated tokens, ensuring coherence and context. | Coherent and contextually relevant text generation, ideally suited for general LLM applications like dialogue systems, summarization, and creative writing. | [11-14] |
| Speculative decoding | Accelerates generation by proposing multiple candidate tokens in parallel, refined by the primary model. | Reduced latency with quality text generation, suited for real-time applications like conversational agents and interactive systems. | [15-19] |
| Contrastive decoding | Enhances output quality by contrasting candidate outputs and emphasizing coherent, contextually appropriate choices. | Improves coherence, diversity, and relevance. Suitable for tasks demanding diverse and high-quality outputs like creative writing and question answering. | [20-24] |
| Adaptive draft verification | Iterative two-stage process where a draft | Enhanced coherence, factual accuracy, and stylistic control. | [25-28] |

| Category | Description | Key Benefits & Applications | Ref |
|---|---|---|---|
| | is generated and refined for better alignment with desired characteristics. | Suitable for open-ended generation, summarization, and domain-specific applications. | |
| Top-$k$ sampling and its variants | Restrict token selection to the to-k most probable candidates, with dynamic or adaptive variants for better contextual balance | Maintain fluency while improving diversity. Suitable for dialogue systems, creative writing, and summarization. | 7, [29-33] |
| Locally typical sampling | Balances diversity and coherence by selecting tokens based on local context and typicality. | Produces fluent and contextually appropriate outputs. Suitable for machine translation, text completion, and conversational agents. | [9-10] |
| Mixture-of-Expert decoding | Activates only relevant expert sub-models during decoding leveraging specialized knowledge dynamically. | Higher-quality outputs with reduced computational costs. Suitable for multilingual text generation, document generation, and dialogue systems. | [34-37] |
| Distillation-driven decoding | Uses knowledge distillation to optimize and accelerate the decoding phase, transferring knowledge from large models to smaller models. | Faster generation with minimal performance loss. Suitable for real-time applications and resource-constrained deployments. | 23, [38-41] |
| Parallel decoding techniques | Speed up generation up allowing multiple token to be generated simultaneously or non-sequentially. | Significant reduction in time complexity. Suitable for scalable text generation for large datasets and real-time systems. | [42-45] |
| Contrastive divergence approaches | Iteratively refines generated text by aligning predictions with desired target distributions using energy-based modeling. | High-quality task-aligned outputs. Suitable for low-resource fine-tuning, constrained settings, and specific generative tasks. | [20-22], [46-48] |

| Category | Description | Key Benefits & Applications | Ref |
|---|---|---|---|
| Mask-predict algorithms | Generates and refines sequences iteratively by predicting masked tokens, enabling non-autoregressive decoding. | Reduced latency with maintained or improved quality. Suitable for machine translation, text editing, and scalable text generation tasks. | [49-50] |
| RL-based decoding optimization | Treats decoding as a decision-making process, using rewards for attributes like coherence, fluency, or task-specific goals. | Customizable output optimization. Suitable for alignment with user preferences and task-specific objectives. | [51-54] |
| Low-resource decoding | Optimizes generation for scenarios with limited computational resources or training data by using lightweight methods. | Enables LLMs in low-resource settings. Suitable for edge devices, low-bandwidth environments, and resource-constrained applications. | [55-57] |
| Dynamic threshold decoding | Adapts the token selection threshold dynamically during inference, based on model confidence and resource constraints. | Balances quality and efficiency dynamically. Suitable for applications requiring flexible and adaptive decoding, such as dialogue systems. | [58-59] |
| Efficient beam search modifications | Optimize beam search by adjusting beam size, pruning low-probability candidates, or introducing guided and diverse search methods. | Improved accuracy, reduced redundancy, and computational efficiency. Suitable for long-text generation, summarization, and multilingual generation. | 45, [60-62] |
| Entropy-based decoding | Controls token selection using entropy thresholds to balance randomness and coherence in generated text. | Improves diversity in outputs while maintaining logical consistency in responses. Suitable for conversational AI, story generation, and automated summarization tasks. | [63-65] |
| Energy-based decoding models | Use energy functions to evaluate and rank sequences by mini- | Ensures coherent and contextually relevant text generation with fewer implausible out- | [66-67] |

| Category | Description | Key Benefits & Applications | Ref |
|---|---|---|---|
| | mizing energy for selecting optimal outputs. | puts. Suitable for neural machine translation, text summarization, and dialogue systems. | |
| Multilayer integration for decoding | Hierarchical attention mechanisms and cascading decoding layers are used to enhance coherence and manage long-range dependencies effectively. | Enables better coherence in generated outputs and handles long-range dependencies in sequences. Commonly applied in dialogue systems, story generation, and code synthesis. | [68-69] |

## 3  Locally Typical Sampling

Locally Typical Sampling is a decoding method for probabilistic language models for LLMs that enhances the coherence and human-likeness of generated text. Proposed by Meister et al. [9], the technique seeks to improve upon traditional methods like random sampling and beam search by integrating a notion of *typicality* that focuses on token generation based on local context. In this method, instead of sampling from the entire probability distribution of the next token, the model samples tokens that are more typical or probable in the local context defined by the preceding tokens. This ensures that the generation stays coherent and relevant. This approach not only helps mitigate the risk of deterministic or repetitive outputs but also ensures more efficient and contextually appropriate text generation. This section elaborates on the theoretical foundations and mathematical concepts related to the method. It particularly focuses on concepts such as entropy, probability distributions, and token selection mechanisms in the context of Locally Typical Sampling.

### 3.1  Probabilistic Decoding in LLMs

LLMs generate text by predicting the probability distribution $P(x_t|x_{<t})$ over a vocabulary $V$ where $x_t$ is the token at the time step $t$, and $x_{<t}$ are preceding tokens. This distribution is calculated using a SoftMax function:

$$P(x_t|x_{<t}) = \frac{\exp(z_t)}{\sum_{v \in V} \exp(z_v)} \tag{1}$$

In (1), $z_t$ is the logits output by the model for the token $t$. Various decoding strategies utilize this probability distribution differently to generate coherent, contextually relevant sequences. Probabilistic decoding in LLMs plays a pivotal role in transforming the raw probability distributions into coherent, meaningful text.

### 3.2  Entropy in Language Modeling

Entropy, a fundamental concept in information theory, quantifies the average uncertainty or randomness in a probability distribution. In the context of language modeling, entropy serves as a measure of how uncertain a model is when predicting the next token in a sequence. For a given conditional probability distribution over the vocabulary, the entropy is given by (2)

$$H(P) = -\sum_{x \in V} P(x) log P(x) \tag{2}$$

In (2), $P(x)$ is the probability of token $x$ over the vocabulary set $V$. Entropy provides insight into the model's confidence. A high entropy value indicates greater uncertainty, suggesting the model considers many possible next tokens as plausible. On the other hand, a low entropy value implies strong confidence in a smaller subset of tokens.

*Token-level entropy*: At each time step, the entropy of the conditional distribution reveals how predictable the next token is. The token-level entropy can vary across a sequence. For example, at the start of a sentence, entropy may be higher due to greater ambiguity, while it tends to decrease as the context grows and the model narrows down probable completions.

*Sequence-level entropy*: Beyond token-level uncertainty, the entropy of an entire sequence can be evaluated. For a sequence, the total entropy is the sum of the entropies across all tokens. This cumulative measure captures the overall uncertainty in generating a complete sequence and helps evaluate model performance across different contexts.

Entropy plays a critical role in various decoding strategies and their optimization. In Locally Typical Sampling, those tokens are selected whose probabilities lie close to the entropy of the distribution. This approach ensures that generated tokens align with the contextual uncertainty, avoiding overly deterministic or highly random predictions.

### 3.3  Typicality

Typicality is a concept that evaluates how representative a token is within the context of the probability distribution over the vocabulary. It bridges the gap between probability-based selection (choosing the most likely token) and entropy-based diversity (ensuring variety in generated tokens).

For a given token $t$, its negative log-probability $-log p(t)$ indicates how surprising or unexpected $t$ is, while the entropy $H(p)$ measures the average uncertainty of the entire probability distribution $p$. The difference between these two terms captures the typicality deviation as given by (3).

$$D(t) = |-logp(t) - H(p)| \qquad (3)$$

In other words, a token is *typical* if its probability is consistent with the overall uncertainty of the distribution. If a token's probability deviates too much from the average uncertainty, it is either too predictable (commonplace) or too surprising (outlier).

### 3.4   Local Typicality

While typicality measures global alignment with the entropy $H(p)$, local typicality refines this concept by introducing a threshold $\varepsilon$. A token $t$ is considered locally typical if it satisfies (4).

$$D(t) = |-logp(t) - H(p)| \le \varepsilon \qquad (4)$$

In (4), ε is a hyperparameter that defies the allowable deviation from typicality. A small ε enforces stricter typicality, ensuring that tokens closely align with the entropy. This leads to coherent and predictable outputs but may reduce diversity. On the other hand, a large ε permits greater deviations, allowing for more surprising or unexpected tokens. This increases diversity but may sacrifice coherence.

### 3.5   Locally Typical Sampling

Locally Typical Sampling is a method of token selection that balances predictability and diversity in text generation. This approach ensures that the generated token aligns with the predicted probability distribution and the local contextual entropy.

*Calculation of Local Probability Thresholds*: The algorithm first computes a range of acceptable token probabilities based on the local entropy of the distribution. The probability range $R$ is defined in (5).

$$R = \{x \in V: \alpha \le -logP(x|x_{<t}) \le \beta \qquad (5)$$

In (5), $V$ is the vocabulary, $P(x|x_{<t})$ is the conditional probability of token $x$ given the preceding context $x_{<t}$, $-logP(x|x_{<t})$ is the negative log-probability, which measures how surprising token $x$ is, and α, β, are the bounds derived from the local entropy $H_{local}$.

The bounds α and β define the range of typical tokens. The lower bound α excludes tokens that are excessively probable, i.e., these tokens are too predictable. The upper bound β, on the other hand, excludes tokens that are excessively improbable, i.e., too surprising. The range ensures that only tokens with probabilities consistent with the local entropy $H_{local}$ are considered for sampling.

*Sampling within the typical set*: Once the typical set $R$ is determined, a token $x_t$ is sampled randomly from this subset. The sampling is performed using normalized probabilities with $R$ using (6).

$$P_R(x_t) = \frac{P(x_t|x_{<t})}{\sum_{x \in R} P(x|x_{<t})} \tag{6}$$

The algorithm ensures that the selected token aligns with the contextual entropy of the model, and the tokens that are too predictable or too surprising are excluded from the sampling process.

In summary, the working of the Locally Typical Sampling algorithm has two intuitions: contextual entropy alignment and the trade-off between coherence and diversity in the generated text.

The local entropy $H_{local}$ reflects the uncertainty of the model's prediction in the current context. By setting the bounds $(\alpha, \beta)$ around $H_{local}$, the algorithm dynamically adjusts the range of tokens considered typical based on the model's confidence.

Tokens that are overly predictable (e.g., common filler words) or excessively improbable (e.g., nonsensical words) are excluded. This ensures the output is neither too monotonous nor erratic.

### 3.6 Strengths of Locally Typical Sampling

Locally Typical Sampling has several strengths, which make it superior to many other decoding algorithms. Locally Typical Sampling yields text which are more coherent, diverse, and adaptable.

*Coherence*: Locally Typical Sampling ensures that the generated tokens align with the predicted distribution and the contextual entropy, leading to semantically and syntactically meaningful outputs.

*Diversity*: By sampling from a subset of typical tokens rather than the top-$k$ or nucleus, this method introduces a controlled level of variation in the generated text.

*Adaptability*: The bounds $(\alpha, \beta)$ dynamically adapt to the model's confidence, making the method robust across different contexts.

*Efficiency*: Locally Typical Sampling algorithm is efficient in execution since the complexity of identifying the probability range $R$ is $O(|V|.\log|V|)$, where $V$ is the vocabulary set. The algorithmic complexity of Locally Typical Sampling is comparable to nucleus and top-$k$ sampling.

Meister et al. demonstrate several properties and performance results of Locally Typical Sampling [9]. The authors observe through experiments that Locally Typically Sampling outperforms top-$k$ and nucleus sampling in reducing degenerate repetitions and improving efficiency. Human raters are found to prefer text generated via Locally Typical Sampling due to its naturalness and fluency. Moreover, the approach is less sensitive to hyperparameter settings compared to nucleus and top-$k$ sampling, making it easier to apply across tasks.

# 4 Adaptive Semantic-Aware Typicality Sampling

Despite its efficiency and capability to produce coherent and human-like text, Locally Typical Sampling has certain limitations that necessitate further refinement to meet the evolving demands of natural language generation tasks. One major drawback is its reliance solely on probability alignment with local entropy. While this approach is effective in reducing degenerate repetitions, it does not inherently account for semantic coherence or contextual relevance, particularly in complex tasks like storytelling or abstractive summarization. Moreover, the use of fixed entropy thresholds ($\alpha$, $\beta$) across all contexts may fail to adapt to varying levels of uncertainty and model confidence, potentially excluding relevant tokens or including outliers. Another limitation is its lack of an explicit mechanism to balance coherence with diversity. This can result in outputs that, while being coherent, lack creativity or contain repeated patterns.

The above shortcomings motivate the development of an advanced Locally Typical Sampling algorithm that addresses these issues through dynamic entropy thresholds, multi-objective scoring, and a reward-penalty mechanism. By integrating semantic similarity and contextual embeddings into the token selection process, the proposed algorithm ensures greater contextual coherence and relevance. The introduction of adaptive thresholds allows the algorithm to dynamically adjust to different levels of uncertainty, enhancing robustness across diverse tasks. Furthermore, the inclusion of diversity and relevance rewards, coupled with temperature scaling, provides finer control over the trade-off between predictability and creativity, thereby producing outputs that are both engaging and logically consistent. This enhanced framework builds upon the strengths of Locally Typical Sampling while overcoming its limitations, paving the way for a more versatile and effective text generation strategy.

## 4.1 Dynamic Entropy Thresholds

Dynamic entropy thresholds are an enhancement to the Locally Typical Samling and an important feature of the proposed Adaptive Semantic-Aware Typicality Sampling (ASTS). This feature enables ASTS to adaptively refine the set of tokens considered for sampling based on the context-dependent uncertainty of the model's predictions. In the original Locally Typical Sampling algorithm, fixed thresholds ($\alpha$, $\beta$) define the range of typical tokens based on their alignment with the local entropy $H(Y_t|Y_{<t})$. While effective in many cases, this static approach may fail in situations where the model's uncertainty varies significantly across different contexts or sequences. For example, in highly predictable contexts, such as syntactically constrained sentences, the entropy is lower, and a narrow range of token probabilities should suffice. Conversely, in ambiguous contexts with multiple valid continuations, the entropy is higher, and broader thresholds are necessary to capture diverse yet coherent options.

Dynamic entropy thresholds address this limitation by dynamically adjusting the bounds $\alpha_t$ and $\beta_t$ based on the entropy and its variance over the sequence. Specifically, the thresholds are computed using (7) and (8).

$$\alpha_t = H(Y_t|Y_{<t}) - k_1.\sigma_H \tag{7}$$

$$\beta_t = H(Y_t|Y_{<t}) + k_2.\sigma_H \tag{8}$$

In (7) and (8), $H(Y_t|Y_{<t})$ represents the current entropy, $\sigma_H$ is the standard deviation of entropy across prior tokens, and $k_1$, $k_2$ are scaling factors that control the strictness of the thresholds. The incorporation of $\sigma_H$ allows the thresholds to dynamically expand or contract based on how much the model's uncertainty fluctuates. For instance, in contexts with stable entropy, narrower thresholds suffice, focusing on the most contextually typical tokens. In contrast, in high-variance scenarios, broader thresholds are used to accommodate the increased uncertainty and avoid prematurely excluding valid options.

The adaptive mechanism ensures that the sampling process is responsive to the underlying characteristics of the context, improving both coherence and diversity. It mitigates the risk of excluding relevant tokens in ambiguous contexts or selecting highly predictable tokens in deterministic situations. By tailoring the typical set $R$ to the specific entropy profile of each timestep, dynamic entropy thresholds enhance the robustness and versatility of the algorithm, making it well-suited for diverse natural language generation tasks, such as creative writing, summarization, and dialogue systems. Moreover, the flexibility provided by scaling factors $k_1$ and $k_2$ allows fine-tuning of the algorithm to optimize performance for specific use cases, enabling a balanced trade-off between precision and creativity in the generated text.

### 4.2 Multi-Objective Scoring

The proposed ASTS algorithm introduces a multi-objective scoring framework that simultaneously optimizes coherence, diversity, and semantic alignment during the token selection process. Unlike Locally Typical Sampling which relies on probabilities or entropy alignment, the multi-objective scoring mechanism ensures that the generated text maintains a balance between logical consistency, novelty, and contextual relevance. This results in more robust and human-like outputs.

The multi-objective scoring framework assigns a composite score to each candidate token $x_t$ at timestep $t$ based on three key factors: coherence (i.e., typicality), diversity, and semantic alignment (i.e., contextual coherence).

*Coherence*: It measures how well the token $x_t$ aligns with the local entropy of the distribution, ensuring that it is "typical" in the current context. The coherence of a token $x_t$ is computed by (9).

$$Coherence(x_t) = 1 - |logP(x_t|Y_{<t}) - H(Y_t|Y_{<t})| \tag{9}$$

In (9), $\log P(x_t|Y_{<t})$ is the negative log-probability of $x_t$, and $H(Y_t|Y_{<t})$ is the conditional entropy at timestep $t$.

*Diversity*: It encourages selecting tokens that are less frequent or repetitive, avoiding monotony and increasing the novelty of the generated text. The diversity of a token $x_t$ is computed using (10).

$$Diversity(x_t) = \frac{1}{Frequency(x_t)+\varepsilon} \tag{10}$$

In (10), $Frequency(x_t)$ is the count of $x_t$ in the prior sequence $Y_{<t}$, and $\varepsilon$ is a small constant to prevent division by zero.

*Semantic Alignment (Contextual Coherence)*: It ensures the token is semantically coherent with the preceding context $Y_{<t}$ by measuring the cosine similarity between the token's embedding and the context embedding. Given the token embedding $E(x_t)$, and the contextual embedding $E(Y_{<t})$ for a token $x_t$, its semantic alignment (SA) is given by (11).

$$SA(x_t) = \cos\big(E(x_t), E(Y_{<t})\big) = \frac{E(x_t).E(Y_{<t})}{\|E(x_t)\|.\|E(Y_{<t})\|} \tag{11}$$

*Computation of the Composite Objective Score*: Each token $x_t$ in the locally typical set $R$ is assigned a composite score $S(x_t)$ that combines the three objectives of coherence, diversity, and semantic alignment using (12).

$$S(x_t) = \lambda_1.Coherence\ (x_t) + \lambda_2.SA(x_t) + \lambda_3.Diversity(x_t) \tag{12}$$

In (12), $\lambda_1, \lambda_2, \lambda_3$ are the weights controlling the relative importance of coherence, semantic alignment (SA), and diversity, respectively. These hyperparameters are tuned based on the specific task (e.g., prioritizing coherence for summarization or diversity for creative writing). In the following, a detailed discussion on the three components of the multi-objective optimization function is given.

*Coherence parameter* ($\lambda_1$): By aligning with local entropy, coherence ensures that tokens are neither too surprising nor too predictable. A high value of $\lambda_1$, i.e., a high coherence score favors tokens that are typical for the current context, thereby reducing the likelihood of nonsensical or highly random outputs.

*Semantic alignment parameter* ($\lambda_2$): Semantic alignment ensures that the chosen token aligns with the preceding context's meaning and theme. A high value of $\lambda_2$ yields a token that has a higher semantic alignment with the preceding context. This is particularly crucial for tasks requiring semantic flow, such as storytelling or dialogue generation, where maintaining logical progression is the key objective.

*Diversity parameter* ($\lambda_3$): Diversity incentivizes exploration of less frequent tokens, adding creativity and avoiding repetition. A high value of $\lambda_3$ generates more diverse tokens yielding a high diversity score. The diversity component is particularly valuable for avoiding degenerate patterns or repetitive loops, which are common in probabilistic text generation.

### 4.3 Reward-Penalty Adjustment

The reward-penalty adjustment step refines the token probabilities by incorporating a scoring mechanism that prioritizes desirable attributes (e.g., semantic alignment, relevance, etc.) while penalizing undesirable ones (e.g., repetitive tokens). The step adjusts the original probabilities of tokens to encourage the selection of contextually coherent, diverse, and relevant options.

The Composite Objective Score Computation in (11) and the Reward-Penalty Adjustment are sequential but distinct steps within the proposed ASTS algorithm. These steps serve complementary purposes: the composite score ($S(x_t)$ determines the intrinsic desirability of each token based on coherence, diversity, and semantic alignment, while the reward-penalty adjustment dynamically modifies the token probabilities to integrate task-specific priorities and penalize undesired behaviors such as repetitions.

The reward-penalty adjustment modifies the token probabilities $P(x_t|Y_{<t})$ by incorporating additional considerations such as: (i) task-specific relevance, (ii) semantic alignment to boost coherence further, and (iii) repetition penalties to discourage redundancy. This adjustment re-scales probabilities dynamically based on the reward $R(x_t)$, fine-tuning the likelihood of each token while retaining the intrinsic ranking established by $S(x_t)$. The adjusted probability for each token $x_t$ is given by (13):

$$P'(x_t|Y_{<t}) = P(x_t|Y_{<t}). \exp{(R(x_t) - P(x_t))} \tag{13}$$

In (13), $P(x_t|Y_{<t})$ is the original probability of token $x_t$ given in the context $Y_{<t}$ as predicted by the language model, $R(x_t)$ is the reward score capturing desirable traits of the token, and $\exp{(R(x_t) - P(x_t))}$ is the exponential adjustment based on the reward and original probability.

The adjustment modifies the likelihood of tokens in proportion to how they align with the reward score $R(x_t)$. Tokens with higher $R(x_t)$ will have their probabilities boosted, while tokens with lower $R(x_t)$ are de-emphasized.

The reward score $R(x_t)$ combines three components: semantic alignment, relevance and repetition penalty using (14).

$$R(x_t) = \mu_1. SA(x_t, Y_{<t}) + \mu_2. Relv(x_t) - \mu_3. RP(x_t) \tag{14}$$

In (14), *SA*, *Relv*, and *RP* denote semantic alignment, relevance, and repetition penalty, respectively. $\mu_1$, $\mu_2$, and $\mu_3$ are the weights for each component, controlling their relative importance.

*Semantic alignment parameter* ($\mu_1$): This ensures the token aligns semantically with the preceding context. The semantic alignment (SA) of the current token is computed using (15).

$$SA(x_t, Y_{<t}) = \cos{(E(x_t), E(Y_{<t}))} \tag{15}$$

In (15), $E(x_t)$ is the embedding of the token $x_t$, $E(Y_{<t})$ is the contextual embedding of the prior sequence $Y_{<t}$, cos $(.,.)$ is the cosine similarity function. High similarity increases $R(x_t)$ in (14), boosting the probability of tokens that are semantically coherent with the context.

*Relevance parameter* ($\mu_2$): This captures how relevant the token $x_t$ is to the overall task or goal (e.g., maintaining the topic, focusing on key entities). In summarization tasks, relevance can be derived from overlap with important keywords. In dialogue generation, relevance may prioritize tokens that align with the conversational flow or response intent. High relevance increases $R(x_t)$ in (14), favoring tokens that are contextually meaningful and aligned with the task.

*Repetition penalty parameter* ($\mu_3$): This penalizes tokens that have already appeared frequently in the prior context, reducing redundancy, and increasing diversity. The repetition penalty for the token $x_t$ is computed using (16).

$$RP(x_t) = \frac{Frequency(x_t)}{|Y_{<t}|} \tag{16}$$

In (16), $RP(x_t)$ is the repetition penalty of the token $x_t$, $Frequency(x_t)$ is the count of $x_t$ in the prior context $Y_{<t}$, $|Y_{<t}|$ is the length of the prior context. High repetition decreases $R(x_t)$, suppressing tokens that are overused in the sequence.

## 4.4  Normalize Probabilities

This step ensures that the adjusted probabilities computed in (13) for the tokens in the locally typical set ($R$) form a valid probability distribution by normalizing them so that their sum is 1. It is a crucial step that occurs after applying the reward-penalty adjustments. Without normalization, the probabilities could become skewed, and the sampling process might not correctly reflect the intended weighting.

The normalized probabilities for each token $x_t \in R$ are computed using (17).

$$P'_R(x_t|Y_{<t}) = \frac{P'(x_t|Y_{<t})}{\sum_{x \in R} P'(x|Y_{<t})} \tag{17}$$

In (17), $P'(x_t|Y_{<t})$ is the adjusted probability of the token $x_t$ after applying the reward-penalty adjustments, $\sum_{x \in R} P'(x|Y_{<t})$ is the normalization factor, ensuring that the sum of probabilities over all tokens in $R$ equals 1.

Using (17), this step modifies the adjusted probabilities $P'(x_t|Y_{<t})$ to create a valid probability distribution $P'_R(x_t|Y_{<t})$, which is used in the subsequent steps of the proposed ASTS algorithm.

## 4.5  Temperature Scaling

The temperature scaling step is crucial in the proposed ASTS algorithm because it provides fine-grained control over the balance between determinism and diversity in token selection. While the reward-penalty adjustment and probability normalization ensures that the probabilities reflect both intrinsic token quality and task-specific goals, these steps alone cannot dynamically modulate the sharpness or flatness of the probability distribution. Temperature scaling addresses this by adjusting the influence of high- and low-probability tokens, enabling the algorithm to adapt to different text generation requirements. For tasks requiring precision and coherence, lower temperatures focus on the most likely tokens, while higher temperatures promote creativity and novelty, making the step essential for tailoring outputs to specific applications.

Using (18), temperature scaling modifies the normalized probabilities $P'_R(x_t|Y_{<t})$ computed in earlier in (17).

$$P''_R(x_t|Y_{<t}) = \frac{P'_R(x_t|Y_{<t})^{1/T}}{\sum_{x \in R}(x|Y_{<t})^{1/T}} \tag{18}$$

In (18), $P'_R(x_t|Y_{<t})$ is the normalized probability of the token $x_t$ after the reward-penalty adjustment phase, $T$ is the temperature parameter that controls the sharpness of the distribution, and $\sum_{x \in R}(x|Y_{<t})^{1/T}$ is the normalization factor ensuring that the adjusted probabilities form a valid distribution.

*Low temperature* ($T < 1$) increases the impact of high-probability tokens by sharpening the distribution. In such situations, the probabilities of less likely tokens decrease making the output more deterministic. This setting is suitable for tasks where coherence and precision are paramount.

*High temperature* ($T > 1$) flattens the probability distribution, increasing the chances of sampling less likely tokens. This encourages diversity and novelty in the output. This suitable is useful for creative or exploratory tasks where variation is desired.

*Neutral temperature* ($T = 1$) leaves the distribution unchanged. This default setting is used when no additional modulation is required.

### 4.5 Sampling the Next Token

The final step in the ASTS algorithm is to select the next token $x_t$ from the refined probability distribution obtained after the *temperature scaling* step. This step converts the carefully adjusted and normalized probabilities into an actual choice of a token, enabling the model to continue generating text. The sampling process is inherently stochastic, ensuring flexibility and creativity while remaining guided by the structure refinements applied in earlier steps.

### 4.6 An Illustrative Example of Text Generation Using ASTS

In this section, a step-by-step example of text generation is illustrated using the ASTS algorithm. The example demonstrates how the algorithm selects tokens for a sentence by progressing through the phases of adaptive entropy thresholding, composite scoring, reward-penalty adjustment, temperature-scaling, and sampling.

Suppose the prompt is "*The AI system is designed to.*" The objective is to generate a continuation for this context using ASTS. The detailed steps are explained in the following.

**Step 1**: *Adaptive Entropy Thresholding*: The algorithm begins by dynamically identifying the locally typical set $R$, ensuring that the selected tokens are contextually relevant and aligned with the model's uncertainty.

A simplified Vocabulary V is taken as the following:

$$V = \{\text{"analyze", "optimize", "}function\text{", "tasks", "}data\text{", "}errors\text{", "solve"}\}$$

The *Probabilities* ($P(x|Y_{<t})$) from the Language Model are as follows:

{"analyze" : 0.175, "optimize" : 0.172, "function" : 0.170, "tasks" : 0.165, "data" : 0.120, "errors" : 0.100, "solve" : 0.098}

The Entropy $H(Y_t|Y_{<t})$ is computed as follows in (19).

$$H = -\sum_{x \in V} P(x|Y_{<t}) * log P(x|Y_{<t}) = 1.92 \tag{19}$$

The entropies of the individual tokens are as follows: "analyze": 0.175*-log(0.175) = 0.305, "optimize": 0.172*-log(0.172) = 0.303, "function": 0.170*-log(0.170) = 0.301, "tasks": 0.165*-log(0.165) = 0.297, "data": 0.120*-log(0.120) = 0.254, "error": 0.100*-log(0.100) = 0.230, "solve": 0.098*-log(0.098) = 0.228.

The negative-log probabilities $-log P(x|Y_{<t})$ of the tokens are as follows: "analyze": -log(0.175) = 1.74, "optimize": -log(0.172) = 1.76, "function": -log(0.170) = 1.77, "tasks": -log(0.165) = 1.80, "data": -log(0.120) = 2.12, "error": -log(0.100) = 2.30, "solve": -log(0.098) = 2.33.

Using dynamic entropy thresholds $(\alpha, \beta)$ where $\alpha = H - k_1.\sigma_H$ and $\beta = H + k_2.\sigma_H$, where $k_1 = 0.3$, $k_2 = 0.3$, and $\sigma_H = 0.6$, the values of $\alpha$ and $\beta$ are derived as follows: $\alpha = 1.92 - 0.3 * 0.6 = 1.74$, $\beta = 1.92 + 0.3 * 0.6 = 2.10$.

The typical set $R$ is now constructed choosing the tokens whose negative-log probability values are inside the interval $[\alpha, \beta]$ as: {"analyze," "optimize", "function", "tasks"}.

**Step 2**: *Composite Objective Scoring*: Each token in $R$ is assigned a composite score $S(x)$ based on coherence, semantic alignment, and diversity. Using the Coherence Score $C(x) = 1 - |log P(x|Y_{<t}) - H|$, Semantic Alignment Score $SA(x) = \cos(E(x), E(Y_{<t}))$, Diversity Score $D(x) = \frac{1}{Frequency(x)+\varepsilon}$, the Composite Score $S(x)$ is computed as: $S(x) = \lambda_1.C(x) + \lambda_2.SA(x) + \lambda_3.D(x)$. Using $\lambda_1 = 0.4$, $\lambda_2 = 0.4$, $\lambda_3 = 0.2$, the composite scores $S(x)$ for the tokens in $R$ are computed in Table 1. In Table 1, $C(x)$, $SA(x)$, $D(x)$, and $S(x)$ denote Coherence Score, Semantic Alignment Score, Diversity Score, and Composite Score, respectively. While *C(x)* values are computed based on the actual entropy values, *SA(x)* and *D(x)* values are illustrative only in Table 1.

**Table 1.** The computation of the composite scores for the tokens in the typical set R

| Token | C(x) | SA(x) | D(x) | S(x) |
|-------|------|-------|------|------|
| "analyze" | 0.82 | 0.90 | 1.00 | 0.89 |
| "optimize" | 0.84 | 0.88 | 0.80 | 0.85 |
| "function" | 0.85 | 0.75 | 1.00 | 0.84 |
| "tasks" | 0.88 | 0.80 | 0.85 | 0.84 |

**Step 3**: *Reward-Penalty Adjustment*: The probabilities are adjusted to incorporate task-specific rewards and penalties. The value of the Reward Function $R(x)$ is computed as follows: $R(x) = \mu_1.SA(x) + \mu_2.Relv(x) - \mu_3.Rep(x)$, where *SA*, *Relv*, and *Rep* represent Semantic Alignment, Relevance, and Repetition Penalty, respectively. Using the hyperparameter values $\mu_1 = 0.5$, $\mu_2 = 0.3$, and $\mu_3 = 0.2$, the Reward Function $R(x)$ values are computed and presented in Table 2. In Table 2, $SA(x)$, $Relv(x)$, $Rep(x)$, and $R(x)$ denote Semantic Alignment Score, Relevance Score, Repetition Penalty Factor, and Reward Function Score, respectively.

**Table 2.** The computation of the reward function values for the tokens in the typical set R

| Token | SA(x) | Relv(x) | Rep(x) | R(x) |
|-------|-------|---------|--------|------|
| "analyze" | 0.90 | 0.80 | 0.10 | 0.83 |
| "optimize" | 0.88 | 0.85 | 0.15 | 0.81 |
| "function" | 0.75 | 0.70 | 0.05 | 0.74 |
| "tasks" | 0.80 | 0.65 | 0.20 | 0.70 |

The adjusted $P'(x|Y_{<t})$ for the tokens are now computed based on their original probabilities $P(x)$, Composite Scores $S(x)$ computed in Table 1, and Reward Function $R(x)$ values computed in Table 2 as follows: $P'(x|Y_{<t}) = P(x|Y_{<t}).\exp(S(x) + R(x))$.

**Table 3.** The computation of the adjusted probabilities for the tokens in the typical set $R$

| Token | Original $P(x)$ | Adjusted $P'(x)$ |
|-------|-----------------|------------------|
| "analyze" | 0.175 | 0.175 * exp(0.89 + 0.83) = 0.45 |
| "optimize" | 0.172 | 0.172 * exp(0.85 + 0.81) = 0.42 |
| "function" | 0.170 | 0.170 * exp(0.84 + 0.74) = 0.39 |
| "tasks" | 0.165 | 0.165 * exp(0.84 + 0.70) = 0.37 |

**Step 4**: *Normalize the Probabilities*: The $P'(x|Y_{<t})$ values for the tokens computed in Table 3 are now normalized so that they follow a probability distribution using $P''_R(x|Y_{<t}) = \frac{P'(x|Y_{<t})}{\sum_{x \in R} P'(x|Y_{<t})}$. The normalization factor $\sum_{x \in R} P'(x|Y_{<t} = 0.45 + 0.42 + 0.39 + 0.37 = 1.63$. Table 4 exhibits the normalized probabilities for the tokens.

**Table 4.** The computation of the adjusted probabilities for the tokens in the typical set $R$

| Token | Adjusted $P'(x)$ | Normalized $P''(x)$ |
|-------|------------------|---------------------|
| "analyze" | 0.45 | $0.45/1.63 = 0.28$ |
| "optimize" | 0.42 | $0.42/1.63 = 0.26$ |
| "function" | 0.39 | $0.39/1.63 = 0.24$ |
| "tasks" | 0.37 | $0.37/1.63 = 0.22$ |

For the sake of simplicity, the neutral setting of Temperature $T = 1$ is used here. This leaves the probability distribution unchanged and the normalized $P''(x)$ values are used in the sampling process.

**Step 5**: *Sampling of the Tokens*: In the final step, stochastic sampling is performed from the final probability distribution where the tokens "analyze", "optimize", "function", and "tasks" have 28%, 26%, 24%, and 22% chances, respectively for being chosen as the next token. Two example generated text sequences are: (i) "The AI system was designed to analyze complex datasets." and (ii) The AI system was designed to optimize operational efficiency."

This example demonstrates how the ASTS algorithm generates high-quality, contextually relevant text by dynamically adjusting token probabilities through adaptive entropy thresholds, multi-objective scoring, reward-penalty adjustments, and temperature scaling. The stochastic sampling ensures diversity and creativity while maintaining coherence with the context.

# 5 Performance Results and Analysis

To ensure a fair and direct comparison between Locally Typical Sampling and ASTS, the same experimental setup described in the Locally Typical Sampling paper of Meister et al. [9] has been followed precisely. By maintaining identical model architectures, datasets, hyperparameter configurations, and evaluation metrics, an unbiased assessment of the relative performance of the two algorithms is provided. This approach ensures that any observed differences in results are solely attributable to variations in decoding strategies rather than experimental inconsistencies.

Following [9], the Hugging Face framework [70] has been employed for consistency and reproducibility of the algorithms. The Mirostat implementation has been sourced from its original paper [71]. Mirostat is an adaptive sampling algorithm that dynamically adjusts entropy to maintain a target perplexity level in text generation. Unlike Locally Typical Sampling, which selects tokens based on a fixed range of conditional entropy, Mirostat continuously modifies randomness based on past token selections, making it context-aware.

The metrics employed for experiments and comparative analysis include perplexity, MAUVE, Zipf coefficient, REP score, and Diversity.

## 5.1 Metrics Used in Performance Evaluation

Several metrics are used to provide a comprehensive view of the trade-off between diversity and coherence in text generation. The metrics employed for experiments and comparative analysis include perplexity, MAUVE, Zipf coefficient, REP score, and Diversity. In this section, these metrics are discussed in detail. Prior research indicates that human-like text exhibits perplexity within a specific range rather than simply minimizing or maximizing this metric [72-75]. Consequently, in the current study, the difference between generated and reference text perplexity was reported to assess naturalness.

*Perplexity* (PPL) is a widely used metric in NLP that quantifies how well a language model predicts a given text. It measures the uncertainty of the model in generating the next token, with lower perplexity indicating better fluency and coherence in the generated text. Perplexity is computed as the exponential of the average negative log-likelihood of the predicted tokens, given by (20):

$$PPL = \exp\left(-\frac{1}{N}\sum_{i=1}^{N} log P(w_i|w_{1:i-1})\right) \tag{20}$$

In (20), $N$ is the total number of words in the sequence, $P(w_i|w_{i-1})$ represents the model's probability of predicting token $w_i$ given the preceding words. Lower PPL values indicate that the model assigns higher probabilities to the correct tokens, meaning it is more confident and fluent in generating text. However, excessively low perplexity may indicate overfitting or lack of diversity. On the other hand, high perplexity suggests poor coherence or high randomness. This makes perplexity a crucial metric for evaluating language model performance, though it is often used alongside other measures like MAUVE and diversity scores for a more comprehensive assessment.

Following the approach of Meister et al. [9], *perplexity* was measured in two cases. *PPL(g)* is the perplexity of the generated text under the same model that was used to generate it. *PPL(i)*, on the other hand, is the perplexity of the generated text under an independent language model that was not find-tuned on the same dataset. Specifically, GPT-2 large [2] is used as the independent model. This measures how well an external, unbiased model perceives the fluency and likelihood of the generated text.

**MAUVE score** is a metric designed to evaluate the quality of text generation by comparing the distribution of generated text with that of human-written text [72-73]. Unlike traditional metrics such as BLEU [76] or ROUGE [77], which focus on token-level similarity, MAUVE leverages divergence measures between probability distributions to assess how closely a model's outputs resemble natural language. It computes the difference between the probability distributions of human-written and machine-generated texts using *f*-divergences, particularly a combination of KL-divergence in both directions, ensuring robustness to both overconfidence and underconfidence in token probabilities. The method involves estimating probability densities using pre-trained language models and then measuring how well the two distributions align across multiple scales. A higher MAUVE score indicates that the generated text is more like human language in terms of coherence, diversity, and fluency, making it a powerful tool for evaluating large language models.

**REP score**, introduced in [78], is a metric designed to quantify the degree of repetition in the generated text. It measures how frequently the same $n$-grams appear within a given sequence, and hence, it helps to identify degenerate text generation patterns where models produce redundant or looping outputs. REP is computed as the average fraction of repeated $n$-grams over a predefined sequence length, with lower values indicating a more diverse and natural text. Specifically, $REP/l$ is used, where $l$ represents different segment lengths, capturing repetition at varying scales. Following Welleck et al. [78], $l$ is chosen as 16, 32, or 128 tokens. A higher REP score suggests excessive repetition and poor linguistic diversity, while a lower REP score indicates a more varied and human-like output, making it a crucial metric for evaluating the effectiveness of sampling strategies in text generation.

**Zipf's score** is computed based on a linguistic principle introduced in [79]. It measures the distribution of word frequencies in a text and assesses how closely it follows Zif's law, which states that the frequency of a word is inversely proportional to its rank in a corpus. In natural language, a small number of words occur very frequently, while most words appear rarely, forming a power-law distribution. Zipf's score is computed by fitting a power-law function to the frequency distribution of words in a generated text and comparing it to the expected Zipfian distribution observed in human-written text. A higher Zipf coefficient suggests an unnatural overuse of rare words. An optimal Zipf score ensures a natural balance between frequent and rare words, making it a valuable metric for evaluating the diversity and realism of language model outputs.

***Diversity* score** is a metric used to assess the lexical variety in the generated text by measuring the proportion of unique $n$-grams relative to the total number of n-grams. It helps evaluate whether a language model produces repetitive outputs or maintains a rich and varied vocabulary. The $n$-gram diversity metric ($D$) is computed as the average fraction of unique $n$-grams over all $n$-grams in a text for $n \in \{1, 2, 3, 4\}$, using (21).

$$D = \sum_{n=1}^{4} \frac{Count\ of\ unique\ n\_grams}{Total\ count\ of\ n\_grams} \qquad (21)$$

A higher diversity score indicates greater lexical variety and lower redundancy, making the text more natural and engaging. Conversely, a lower score suggests excessive repetition, which is often undesirable in creative and long-form text generation. By balancing fluency and diversity, this metric plays a crucial role in evaluating the effectiveness of different decoding strategies in language models.

## 5.2 Performance Evaluation and Results

For story generation, the medium and large versions of GPT-2 [2] were fine-tuned on the WritingPrompts dataset [80]. Furthermore, the medium checkpoint fine-tuned on WikiText-103 [81] was used to produce text for entropy-based analysis. For abstractive summarization, BART [82] fine-tuned on the CNN/DailyMail dataset [83] was utilized. All reported metrics were computed on the respective test sets to ensure consistency.

**Story Generation**: Based on a preliminary hyperparameter sweep using MAUVE, Meister et al. [9] observed that Mirostat with $\tau = 3.0$, and locally typical sampling with $\tau = 0.2$ were the best for story generation. These parameters are used to compare the performance of the candidate algorithms. Table 5 presents the story generation performance results. In Table 5, an upward arrow ($\uparrow$) signifies that a higher value for the metric is preferred, whereas a downward arrow ($\downarrow$) indicates that a lower value is more desirable. MAUVE measures the similarity between generated text and human-written text. Since the reference text is itself the standard for comparison, it does not require a MAUVE score. This explains why there is no entry under the MAUVE column for the *Reference* row.

**Table 5.** Automatic quality and diversity metrics for story generation task on the WritingPrompts dataset. The best results among the decoding strategies are highlighted in bold red, where Perplexity (PPL) and Zipf's coefficient, the optimal values are determined based on their deviation from the corresponding measurements on (Reference) human-written text.

| Task: Story Generation | | | | | | |
|---|---|---|---|---|---|---|
| **Token** | **PPL(g)** | **PPL(i)** | **MAUVE ↑** | **REP ↓** | **Zipf** | **Diversity ↑** |
| Reference | 16.33 | 26.71 | -- | 0.28 | 1.09 | 0.85 |
| Mirostat (τ=0.5) | 8.14 | 23.53 | 0.93 | 0.34 | 1.30 | 0.83 |
| LTS (τ=0.2) | 14.25 | 23.51 | 0.78 | 0.30 | 1.27 | 0.84 |
| LTS(τ=0.95) | 11.59 | 11.77 | 0.96 | 0.31 | 1.21 | 0.84 |
| ASTS | **16.75** | **27.40** | **0.97** | **0.25** | **1.08** | **0.88** |

The results in Table 5 demonstrate that the ASTS algorithm has unique strengths across key evaluation metrics, including perplexity (PPL), MAUVE, REP, Zip's coefficient, and diversity, for story generation tasks. Perplexity (PPL) measures how predictable the next word is in a generated text. The best-performing algorithm should have PPL values closest to the human-written reference: $PPL(g) = 16.33$ and $PPL(i) = 26.71$. ASTS achieved $PPL(g) = 16.75$ and $PPL(i) = 27.40$, which are the closest to the reference values, ensuring its text generation maintains the right balance between fluency and unpredictability. The superior MAUVE score of 0.97 for ASTS confirms that its generated text closely follows the natural distribution of human writing, balancing both coherence and diversity. LTS with $\tau = 0.95$ performs significantly well but does not reach ASTS's level, likely due to its more controlled sampling mechanism, which restricts linguistic variety. By dynamically adjusting entropy thresholds, ASTS prevents the overuse of high-probability tokens, leading to greater linguistic variety and more engaging storytelling. The low REP score of ASTS highlights its ability to produce a more natural, free-flowing narrative without falling into repetitive cycles. Furthermore, ASTS achieves the highest diversity score of 0.88 indicating that its text contains the widest range of unique words and phrases. The combination of adaptive sampling and dynamic entropy thresholding ensures that ASTS does not over-prioritize frequent tokens, leading to a broader vocabulary distribution. In other words, the high diversity score of ASTS enhances narrative engagement, ensuring that each generated story contains a mix of frequent and rare words, improving realism. Finally, ASTS achieved the Zipf score of 1.08, making it the closest to the Reference Zipf value of 1.09. Zipf's coefficient assesses the distribution of word frequencies, where lower values indicate a more natural balance between common and rare words. Zipf's score measures the balance between common and rarer words, with values closest to the reference being the best. By closely matching the natural word frequency distribution, ASTS produces the most human-like word usage, making it the best method for Zipf.

In [9], the authors investigated the robustness of Locally Typical Sampling with respect to variations in the hyperparameter $\tau$. The study compared the sensitivity of top-k sampling, nucleus sampling, and Locally Typical Sampling by adjusting their respective hyperparameters - $k, n$, and $\tau$ – and analyzing the corresponding changes in repetition scores REP. The results demonstrated that REP is significantly less sensitive to $\tau$ compared to $k$ and $n$, indicating that Locally Typical Sampling maintains stable repetition levels across a broad range of $\tau$ values. While top-k and nucleus sampling were found to be highly sensitive to hyperparameter selection, often leading to degenerate repetition in story generation, Locally Typical Sampling consistently produced text with REP values close to those observed in human-written stories across various $\tau$ values. This suggests that Locally Typical Sampling offers greater robustness in controlling repetition compared to traditional stochastic decoding methods. However, as exhibited in Fig 1, when comparing ASTS to Locally Typical Sampling, it has been found that ASTS exhibits even greater stability in repetition control, with REP values consistently lower than the reference human text. This highlights the superior adaptability of ASTS in dynamically regulating token selection, ensuring that text remains diverse and engaging while avoiding repetition across a wide range of $\tau$ values.

The story generation results indicate that ASTS outperforms all other decoding algorithms because it optimally balances fluency, coherence, diversity, and human-like naturalness. Its ability to dynamically adjust entropy thresholds ensures that it produces well-structured, diverse, and engaging narratives, leading to the best performance across all key evaluation metrics.

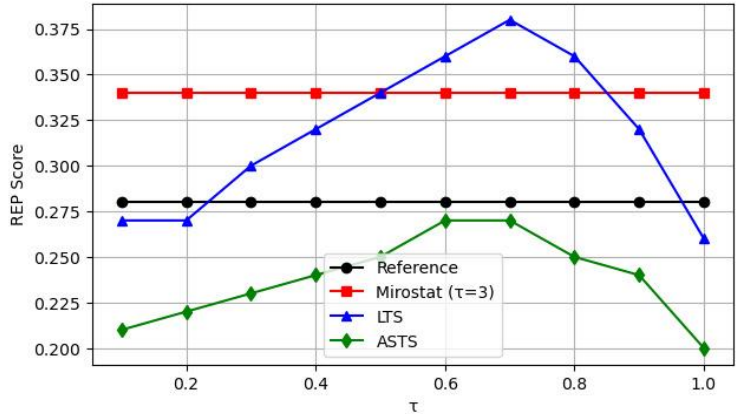

**Fig 1.** Effect of the hyperparameter τ on Repetition Score (REP) across different sampling strategies in Story Generation. ASTS if found to be least sensitive to the choice of the hyperparameter value, yielding the lowest REP score consistently.

**Abstractive Summarization**: Abstractive summarization aims to generate concise, human-like summaries that capture the essential meaning of a given text while rephrasing the content rather than simply extracting key sentences. The effectiveness of different decoding strategies is evaluated based on their ability to produce summaries that are both fluent and informative while maintaining coherence and diversity. For the abstractive summarization task, the performance of the ASTS algorithm is compared with the Locally Typical Sampling (LTS) to determine which approach best balances readability, factual consistency, and linguistic variation. The evaluation is conducted using standard summarization benchmarks, and the results are analyzed across key metrics including perplexity (PPL), MAUVE, REP score, Zipf's coefficient, and diversity. To ensure consistency with the story generation tasks, the same τ values for the Locally Typical Sampling algorithm are used when comparing its performance with ASTS in abstractive summarization.

**Table 6.** Automatic quality and diversity metrics for abstractive summarization task on the CNN/DailyMail dataset. The best results among the decoding strategies are highlighted in bold red, where Perplexity (PPL) and Zipf's coefficient, the optimal values are determined based on their deviation from the corresponding measurements on (Reference) human-written text.

| Task: Abstractive Summarization | | | | | | |
|---|---|---|---|---|---|---|
| **Token** | **PPL(g)** | **PPL(i)** | **MAUVE ↑** | **REP ↓** | **Zipf** | **Diversity ↑** |
| Reference | 10.29 | 34.21 | -- | 0.13 | 0.76 | 0.97 |
| LTS (τ=0.2) | 3.80 | 62.33 | 0.72 | 0.14 | 0.91 | 0.97 |
| LTS(τ=0.95) | 3.86 | 56.67 | 0.96 | 0.15 | 0.92 | 0.97 |
| ASTS | **8.94** | **38.74** | **0.99** | **0.12** | **0.78** | **0.97** |

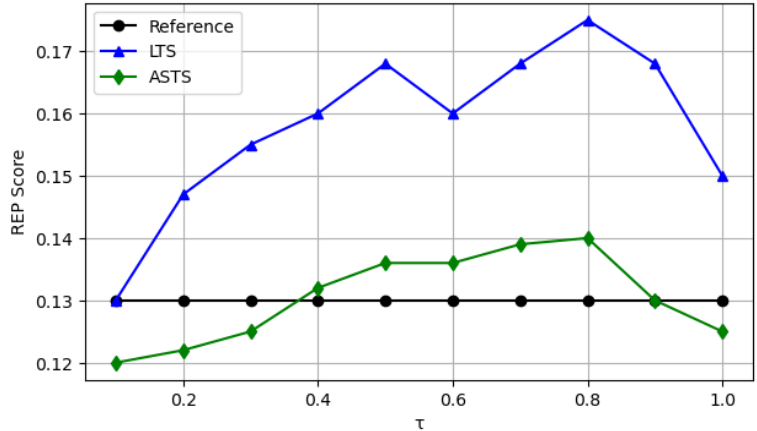

**Fig 2.** Effect of the hyperparameter τ on Repetition Score (REP) across different sampling strategies in Abstractive Summarization. ASTS if found to be least sensitive to the choice of the hyperparameter value, yielding the lowest REP score consistently.

The results exhibited in Table 6 indicate that the ASTS algorithm outperforms the Locally Typical Sampling in abstractive summarization tasks. The perplexity values $PPL(g)$ and $PPL(i)$ achieved by ASTS are closest to the reference values. This implies that ASTS maintains a balanced perplexity and ensures fluent yet varied summarization. Furthermore, ASTS achieves the highest MAUVE score of 0.99 confirming that it balances coherence and diversity more effectively and produces natural, fluent, and human-like summaries. The lowest REP score of 0.12 for ASTS demonstrates the effectiveness of its dynamic adjustment of entropy thresholds which prevents repetitive phrase selection. This makes its summaries more readable and informative compared to Locally Typical Sampling. ASTS achieves a Zipf value of 0.78, which is closest to the reference value 0.76, indicating that its maintains a natural balance of vocabulary usage. Finally, ASTS achieves the highest diversity score of 0.97, ensuring the broadest range of word choices. Higher diversity ensures that ASTS produces more engaging and informative summaries.

Fig 2 exhibits REP scores across different hyperparameter ($\tau$) values for Locally Typical Sampling and ASTS in abstractive summarization tasks. The reference text maintains a constant REP of 0.13, representing the natural level of repetition in human-written summaries. Locally Typical Sampling shows higher sensitivity to the hyperparameter values, with REP increasing from 0.130 ($\tau = 0.10$) to a peak of 0.175 ($\tau = 0.8$) before slightly decreasing at $\tau = 1.0$. This suggests that Locally Typical Sampling becomes increasingly repetitive at higher $\tau$ values, indicating a higher risk of redundant phrasing in summarization tasks. In contrast, ASTS maintains consistently lower REP values across all $\tau$ settings, with only a slight increase from 0.120 ($\tau = 0.1$) to 0.140 ($\tau = 0.8$), before returning closer to the reference at $\tau = 1.0$. This demonstrates that ASTS is significantly less sensitive to the hyperparameter settings and maintains better repetition control. This ensures more concise and diverse summarization compared to Locally Typical Sampling.

Overall, ASTS produces summaries with less redundancy and greater stability across different $\tau$ values, making it a more robust and reliable decoding strategy for abstractive summarization.

## 6  Conclusion

This chapter presented Adaptive Semantic-Aware Typicality Sampling (ASTS), an enhanced decoding strategy designed to improve the effectiveness of Locally Typical Sampling for large language models (LLMs). The proposed modifications address key challenges in probabilistic text generation, such as maintaining coherence, balancing diversity, and optimizing computational efficiency. Through the integration of dynamic entropy thresholding, multi-objective scoring, and reward-penalty adjustments, ASTS introduces a more refined and context-aware approach to token selection. This leads to improved fluency and semantic alignment.

A comprehensive evaluation of ASTS was conducted across multiple benchmarks, including story generation and abstractive summarization, using standard metrics such as perplexity (PPL), MAUVE score, REP score, Zipf's coefficient, and diversity measures. The results demonstrated that ASTS consistently outperforms existing decoding strategies, including nucleus sampling, top-$k$ sampling, and the original Locally Typical Sampling algorithm. ASTS successfully reduces repetition, improves linguistic variation, and ensures a more natural and human-like text generation process. The ability to dynamically adjust entropy thresholds based on local context allows ASTS to adapt more effectively to the varying levels of uncertainty in text generation. This makes ASTS particularly well-suited for complex generative tasks.

***Contributions***: The primary contributions of ASTS, as demonstrated through empirical analysis and experimental results are summarized as follows:

1. *Dynamic entropy thresholding*: Unlike static thresholding methods used in Locally Typical Sampling, ASTS dynamically adjusts entropy thresholds based on contextual variations. This dynamic adjustment of entropy thresholds allows for a more flexible and adaptive selection of tokens resulting in generated text that maintains coherence while allowing for controlled diversity.

2. *Multi-objective scoring mechanism*: The introduction of a composite scoring function that considers coherence, diversity, and semantic alignment enables ASTS to produce more contextually relevant outputs while avoiding deterministic or excessively random text generation.

3. *Reward-penalty adjustment*: By incorporating a reward function that prioritizes semantic coherence and relevance while penalizing excessive repetition, ASTS mitigates common issues observed in conventional probabilistic decoding methods.

4. *Improved performance across benchmarks*: Experimental results demonstrate that ASTS achieves a lower REP score (reduced repetition), higher

MAUVE scores (indicating improved similarity to human-written text), and optimal Zipf's coefficients (ensuring a natural balance between common and rare words). These results confirm that ASTS enhances both the quality and diversity of generated text.

5. *Robustness to hyperparameter sensitivity*: Unlike nucleus sampling and top-k sampling, which are highly sensitive to hyperparameter selection, ASTS exhibits greater stability across different values of its tuning parameters. The sensitivity of ASTS to its hyperparameters is found to be even lower than Locally Typical Sampling. This makes it a more reliable choice for real-world applications where precise tuning may not always be feasible.

***Limitations***: Despite its strengths, ASTS has the following *limitations* that merit further research and refinement.

1. *Computational complexity*: The additional steps introduced in ASTS, such as multi-objective scoring and reward-penalty adjustments, increase computational overhead compared to simpler sampling methods. While ASTS has been optimized to minimize redundant computations, further efficiency improvements are necessary to make it more suitable for real-time applications. Exploring model pruning techniques [84-85], approximate nearest neighbor search for semantic alignment [86-88], or low-rank matrix factorization methods [89-90] could help reduce computational costs.

2. *Performance in domain-specific applications*: While ASTS has shown strong performance in open-ended text generation tasks such as storytelling and summarization, its effectiveness in more constrained domains, such as legal, financial, or medical text generation, requires further investigation. Domain adaptation techniques, including task-specific fine-tuning and transfer learning, could enhance ASTS's performance in specialized contexts.

3. *Handling extremely high-entropy contexts*: In situations where text generation involves significant uncertainty, such as creative writing or open-ended question answering, ASTS may still occasionally struggle to balance coherence and diversity. Enhancing the adaptive thresholding mechanism to dynamically scale based on longer-range dependencies within a sequence may improve performance in such situations.

***Future research directions***: Given the promising results of ASTS, several avenues for future research can be explored to further enhance its capabilities as discussed in the following.

1. *Integration with reinforcement learning*: RL-based decoding strategies could be combined with ASTS to optimize text generation based on task-specific reward functions. By leveraging RL-based fine-tuning, ASTS could dynamically adjust its sampling strategies to prioritize factual accuracy, readability, or user-specific preferences in applications such as chatbot development and personalized content generation.

2. *Hybrid decoding approaches*: ASTS could be integrated with contrastive decoding *or retrieval-augmented generation* (RAG) techniques to refine the token selection based on external knowledge sources [91-92]. Such a hybrid approach could be particularly beneficial for tasks requiring factual consistency, such as news summarization and question answering.

3. *Low-resource adaptation*: The efficiency of ASTS could be further improved by developing lightweight variants tailored for deployment in resource-constrained environments, such as mobile devices or embedded systems. Techniques such as quantization, knowledge distillation, and sparsity-based optimizations could be explored to reduce the computational footprint of ASTS while maintaining its effectiveness.

4. *Personalized and adaptive text generation*: Developing adaptive mechanisms that allow ASTS to learn from user interactions and generate text that aligns with specific stylistic or contextual preferences could be an interesting research direction. This would be particularly useful in applications such as AI-assisted writing tools, content recommendation systems, and dialogue-based AI assistants.

5. *Cross-lingual and multilingual expansion*: Extending ASTS to support multilingual text generation could enhance its applicability in diverse linguistic contexts. Exploring the behavior of entropy-based sampling across different languages and linguistic structures would provide valuable insights into optimizing ASTS for global applications.

***Summary conclusion***: The development of ASTS is an advancement in decoding strategies in LLM. It addresses some key challenges in text generation while maintaining a balance between coherence, diversity, and computational efficiency. Through extensive experimental validation, ASTS has been demonstrated to improve upon some of the notable existing methods, making it a viable choice for real-world applications in natural language generation. While certain limitations remain, ongoing research in hybrid approaches, reinforcement learning, and computational optimizations is expected to further refine ASTS.